\documentclass[pdflatex,sn-mathphys-num]{sn-jnl}

\usepackage{makecell}
\usepackage{graphicx}%
\usepackage{multirow}%
\usepackage{amsmath,amssymb,amsfonts}%
\usepackage{amsthm}%
\usepackage{mathrsfs}%
\usepackage[title]{appendix}%
\usepackage{xcolor}%
\usepackage{textcomp}%
\usepackage{manyfoot}%
\usepackage{booktabs}%
\usepackage{algorithm}%
\usepackage{algorithmicx}%
\usepackage{algpseudocode}%
\usepackage{listings}%
\usepackage{csquotes} 

\theoremstyle{thmstyleone}%
\theoremstyle{thmstyletwo}%

\theoremstyle{thmstylethree}%

\begin{document}

\title[Article Title]{RAISE: Reasoning Agent for Interactive SQL Exploration}


\author[1]{\fnm{Fernando} \sur{Granado}}\email{f171517@dac.unicamp.com}

\author[1]{\fnm{Roberto} \sur{Lotufo}}\email{lotufo@unicamp.com}

\author[1]{\fnm{Jayr} \sur{Pereira}}\email{jayr@unicamp.com}

\affil[1]{\orgdiv{Universidade Estadual de Campinas (UNICAMP)}}


\abstract{Recent advances in large language models (LLMs) have propelled research in natural language interfaces to databases. However, most state-of-the-art text-to-SQL systems still depend on complex, multi-stage pipelines. This work proposes a novel agentic framework that unifies schema linking, query generation, and iterative refinement within a single, end-to-end component. By leveraging the intrinsic reasoning abilities of LLMs, our method emulates how humans answer questions when working with unfamiliar databases: understanding the data by formulating hypotheses, running dynamic queries to validate them, reasoning over the results, and revising outputs based on observed results. Crucially, our approach introduces a new strategy for scaling test-time computation in text-to-SQL: we scale the depth of interactive database exploration and reflection. This shift enables the model to allocate computation dynamically to better understand the data, especially useful in ambiguous and underspecified scenarios. Our experiments show that it improved the Execution Accuracy (EX) from 44.8\% to 56.5\% on the challenging BIRD dataset using \textsc{DeepSeek-R1-Distill-Llama-70B}. Furthermore, when equipped with steps to add more diversity to the answers, our agent achieves a Best-of-N accuracy of 81.8\% with 8 rounds of candidate generation, rivaling the 82.79\% achieved by the top-ranked published solution, while reducing engineering complexity. These findings position our unified framework as a promising alternative for building natural language interfaces to databases.}




\maketitle

\section{Introduction}\label{sec1}


Answering natural language questions using information stored in databases is a critical task across many sectors of the economy, playing a central role in data-driven decision-making. The recent advances in large language models (LLMs) have opened new possibilities in automating the text-to-SQL task, as these models exhibit strong reasoning abilities and extensive common-sense knowledge \cite{brown2020language,bubeck2023sparks,wei2022chain,bib4,bib5,bib6,bib7}. However, even integrating LLMs into sophisticated pipelines, recent benchmarks still reveal a performance gap between machines and humans in this domain \cite{bib5,bib1,bib2,bib3}.

Contemporary text-to-SQL systems often rely on complex multi-component pipelines, delegating tasks such as schema linking, query structuring, and result validation to specialized modules \cite{bib6,bib8,dong2023c3,wang2023mac,bib1,bib2,bib3}. In addition, despite the successful use of test-time compute in the broader AI landscape \cite{jaech2024openai,snell2024scaling}, current pipelines mostly restrict their application to iterative query refinement and query selection strategies such as self-consistency or best-of-N \cite{bib2,bib3, bib8,cao2024rsl,lee2024mcs}. Humans, on the other hand, often spend significant time exploring the database to better understand what data is available, how the data is stored, interpret ambiguous column names, and identify inconsistencies in the data. These systems, however, do not scale compute to support such behavior. 

Although modular pipelines can enhance accuracy, they also increase engineering complexity. Given the reasoning and generalization capabilities of LLMs, an important question arises: Can we simplify these pipelines by allowing the model to handle more of these tasks autonomously? Specifically, can an end-to-end agent dynamically reason about schema linking, formulate queries, iterate over results, identify and handle ambiguities, and allocate more time to these steps when needed, all without extensive manual design?

Aiming to answer this question, this paper presents a novel agentic approach that unifies schema linking, query composition, and query refinement in a single component. Our approach leverages the model’s internal reasoning traces to guide interactions with the database—exploring tables, interpreting results, and revising hypotheses in a manner reminiscent of human analysts. This introduces a new strategy for scaling test-time compute: increasing database exploration and reflection on query results to improve the model's understanding. Without any fine-tuning, our agent is capable of generating the correct solution in 81.8\% of the questions on the BIRD dataset \cite{li2024can}, rivaling the performance of leading multi-stage pipelines for candidate generation, such as the 82.79\% achieved by the start-of-the-art end-to-end solution \cite{bib2}. This challenges the prevailing view that LLMs require extensive external support to perform competitively on text-to-SQL tasks.

The main contributions of this paper are:

\begin{enumerate}[1.]

\item A unified agentic framework that performs schema linking, query generation, and refinement within a single LLM-based component\footnote{All prompts and a sample of the generated data are available in this\href{https://www.dropbox.com/scl/fo/5juy87vmha7w05dreocxw/AEB4hrJbq-RUixdmvGKNDj8?rlkey=dnmxl8d7bibua81jl1631b9me&st=96ej2a2f&dl=0}{Dropbox folder}.}; 

\item A new strategy of scaling test-time computation in Text-to-SQL: deeper database exploration through increased model interactions;

\end{enumerate}

The remainder of this paper is structured as follows: Section~\ref{related_work} reviews related work in modular pipelines for text-to-SQL. Section~\ref{methodology} introduces our proposed agentic framework, describing its reasoning process, database interaction tools, control mechanisms, and strategies for generating diverse outputs. Section~\ref{exp_setup} then presents our experimental setup, including the evaluation framework and controlled comparisons between agents. Results are analyzed in Section~\ref{results}, highlighting the impact of dynamic exploration and pipeline enhancements. Finally, Section~\ref{conclusion} concludes the paper.

\section{Related Work}\label{related_work}

Modern text-to-SQL solutions rely on a complex arrangement of independent components assembled to generate the final SQL query \cite{bib6,bib1,bib2,bib3, dong2023c3,pourreza2023din,wang2023mac,lee2024mcs,xie2025opensearch,pourreza2025reasoning}. Some of the most important components are: 1) the schema linking, responsible for filtering the relevant tables and columns to answer the question; 2) a generation module, responsible for generating many high-quality and diverse set of candidate solutions; and 3) a selection module, responsible for selecting the most appropriate SQL among the generated candidates. This architecture underpins some of the best-performing solutions on the BIRD benchmark\footnote{\url{https://bird-bench.github.io/}}. 

CHESS \cite{bib1} introduced a sophisticated schema linking module that inspired many following works \cite{bib2,bib3,xie2025opensearch}. It follows a multi-stage pipeline that begins with preprocessing and proceeds through a three-step funnel that incrementally narrows down the relevant schema elements. During preprocessing, CHESS detects entities mentioned in the input question and uses Locality-Sensitive Hashing \cite{datar2004locality} and edit distance to match them to similar column values. Embedding-based similarity is used to retrieve relevant descriptions from the database catalog, and heuristics, such as always including foreign keys, ensure the inclusion of essential schema elements. These outputs condition the prompts used by each stage of the funnel.

The generation phase is often implemented as an iterative process: candidate queries are generated, executed, and refined based on the observed results until satisfactory answers are reached. This is also the main place where test-time compute is scaled by increasing the number of candidates and refining steps. CHASE-SQL \cite{bib2} and XIYAN-SQL \cite{bib3} emphasize the importance of generating a wide range of diverse candidates to increase the likelihood of a correct answer among them. CHASE-SQL introduces three independent generation agents, using different strategies to construct the query. While XIYAN-SQL fine-tunes the model in two steps to improve the model's generation capabilities and alignment with the syntactic preferences of the benchmark. At inference time, it selects training examples similar to the input question for inclusion in the prompt to encourage diverse outputs. IAD \cite{bib8} uses a verifier agent to provide feedback to the generator, improving the quality of each candidate.

From the resulting pool of candidates, the final query is selected based on execution outcomes. Some approaches, such as \cite{bib2} and \cite{bib3}, train specialized models specifically to evaluate and select the best-performing query.

In summary, the current top-performing pipelines rely on extensive human engi-
neering to handle the complexity of the problem and provide limited space for the
models to control and reason over the whole available data.

\section{RAISE: Reasoning Agent for Interactive SQL
Exploration}\label{methodology}

\begin{figure}[h]
    \centering
    \includegraphics[width=1\textwidth]{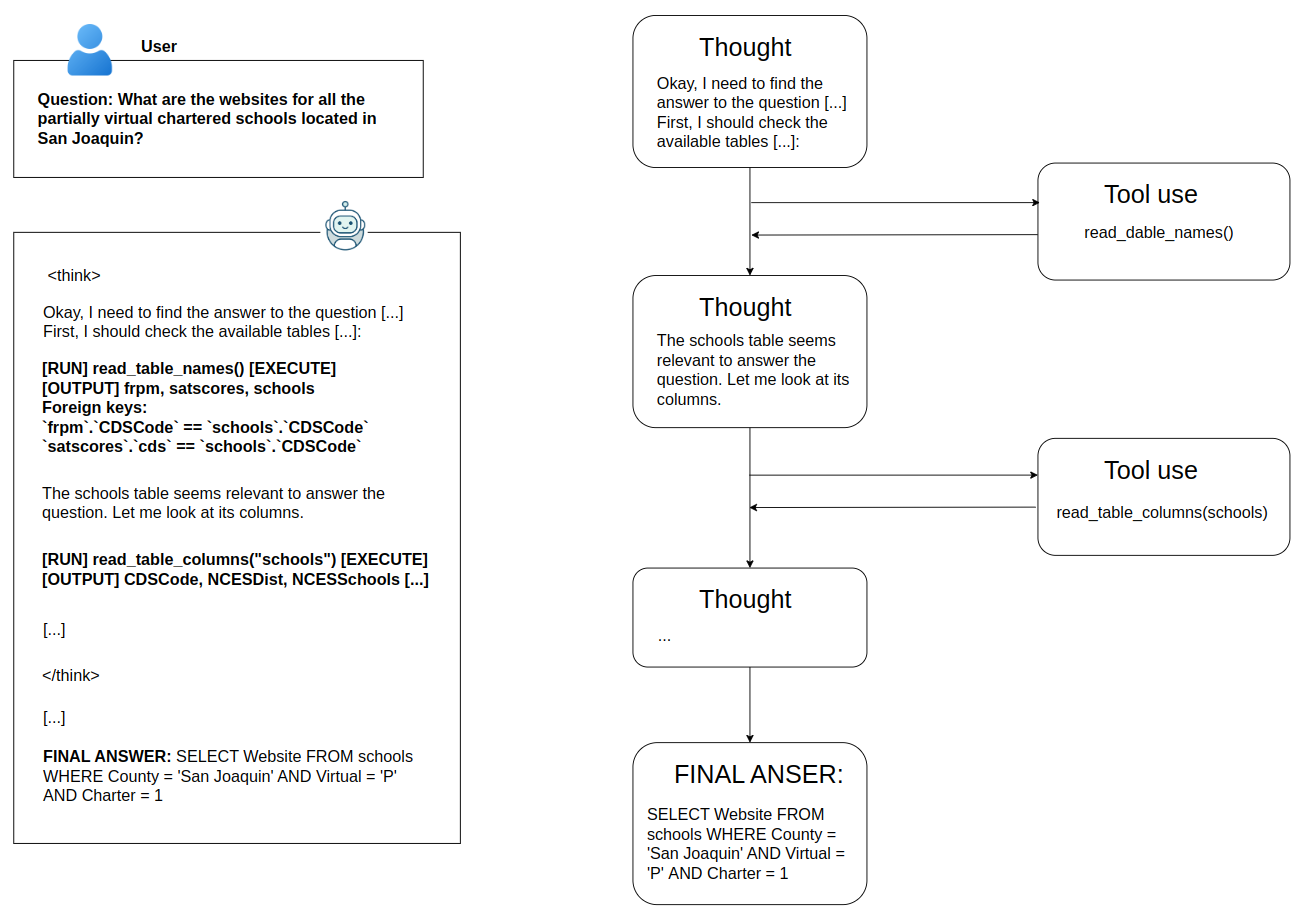}
    \caption{Representation of the interactive and selective database exploration, where the model explores its data and reasons over it to return the final solution.}
    \label{step_page}
\end{figure}

We propose an end-to-end agent designed to answer natural language questions over relational databases by leveraging the reasoning capabilities of LLMs in conjunction with a set of structured tools. Unlike traditional systems that rely on rigid, multi-stage pipelines, our agent has the flexibility to explore the database and solve questions, similar to how human analysts do. The model can form hypotheses, query the data, reflect on intermediate results, and iterate until reaching a confident answer. Figure~\ref{step_page} provides a visual representation of this agent.

\subsection{Thought-Based Planning}\label{thought_based_planning}

At the core of our method is a reasoning LLM, which explicitly generates thoughts before producing the final answer. These models are capable of handling complex, multi-step reasoning tasks, where they can break down intricate problems into manageable components, identify potential errors, and iteratively refine intermediate steps to reach accurate solutions \cite{jaech2024openai}. Recent studies have demonstrated that such models, including Deep Research-style architectures\footnote{\url{https://openai.com/index/introducing-deep-research/}}, excel in structured planning and self-correction, providing robust performance across diverse, high-complexity contexts. In our experiments, we used DeepSeek-R1-Distill-Llama-70B \cite{guo2025deepseek} to implement this core role.

To support interaction with the database, we extend the agent’s capabilities through tool use. In our framework, tools provide access to both database content and schema documentation. The agent is equipped with the following four tools:

\begin{itemize}[label=-] 

\item read\_table\_names()
\item read\_table\_columns(table\_name: str)
\item read\_columns\_documentation(column\_names: list[str])
\item run\_query(sql: str)

\end{itemize}

These four tools are designed to replicate the actions a human would take to efficiently gather the information needed to answer questions. This includes identifying relevant tables and columns, formulating and validating assumptions about data and how it is stored. For instance, the agent might run arbitrary queries to check if any record in a table is missing from another; verify if column values consistently follow a specific format; refine their understanding of an ID column by determining what data holds unique per ID; or cross-check the consistency between the database documentation and actual values. At test time, the agent can scale these actions to run more checks and collect more information, enabling precise comprehension of the database. Moreover, this lets it explore only the relevant portions
of the database, helping to keep the context window compact. As a result, the framework provides full access to the available data while maintaining the context window within manageable limits.

To operationalize this interaction, we define a special tag, \texttt{[EXECUTE]}, which signals when the model has issued a command to be executed. Upon detecting this tag, we parse the tool invocation from the generated text, execute it using a Python script, and append the result back into the model’s context. Token generation then resumes from this updated state, enabling the model to iteratively reason, observe outcomes, and take further actions in a closed feedback loop.

\subsection{Control}\label{control}

We observed that, when used as-is, the model occasionally exhibits undesirable behaviors in certain scenarios:

\begin{enumerate}[1.]
\item When simply given the prompt, the model tends to answer the question too directly, without sufficiently understanding the available data. In many cases, it skips exploring parts of the database that are clearly relevant and would significantly alter its final response if examined.

\item When encountering ambiguous results -- such as queries returning zero rows -- or even in the absence of any clear issue, the model may become stuck in lengthy and sometimes repetitive reasoning loops that fail to make meaningful progress toward a solution.

\end{enumerate}

To improve the model’s behavior, encouraging more thoughtful exploration of the database, steady progress toward a solution, and the consistent production of final answers, we introduced the following adjustments:

\begin{enumerate}[1.]
\item We hardcoded the beginning of every reasoning trajectory with the phrase: \textquote{\textit{Before thinking about the solution, I will have a deep understanding of the data and not make any assumptions about it.}} This helps steer the model toward data exploration before attempting to answer.

\item We implemented a cap on the number of tokens the model can generate without executing any operation. If more than 1400 tokens are used without a tool call, we forcibly append the phrase: \textquote{\textit{Wait, I am thinking for too long without interacting with the database. I can run queries and see the results with the command [RUN] run\_query(...) [EXECUTE]}} in a new line.

\item If the model fails to produce a final answer within 10,000 tokens, we force termination by appending: \textquote{\textit{I am thinking for too long. I will generate my final solution now.}}.

\end{enumerate}

\subsection{Generating Diverse Answers}\label{diversity_generation}

Generating a diverse set of candidate solutions is important to increase the chances of finding the correct query, which can be selected by a posterior module. Moreover, the BIRD benchmark contains a series of implicit preferences that makes the evaluation of queries wrong even in cases where it follows the right logic. For instance, the result is evaluated as wrong if columns are selected in a different order or if an
extra column is included; the gold queries often join all tables before filtering, which may lose records in case records in a table are missing in the other, among many other preferences. By generating a diverse set of candidates, we can increase the chances of having a candidate evaluated as correct. A selection model fine-tuned to select the correct preferences can, then, select one of them.

We notice that increasing the temperature is not enough to increase the diversity of the answers, as each model has very strong biases. However, we find that different models have different biases, which is an effective way to increase diversity. Therefore, after the
initial model completes its exploration, including all executed operations and their
results, we prompt two different LLMs, Claude 3.7 Sonnet and o3-mini, to read this exploration and generate the final
answer independently. We also post-process solutions, changing only the selected columns,
without changing the query logic. This is done by querying a model to specifically identify what columns should be selected and in what order. The full RAISE pipeline is shown in Figure~\ref{full_pipeline}

\section{Experimental Setup}\label{exp_setup}

\begin{figure}[h]
    \centering
    \includegraphics[width=1\textwidth]{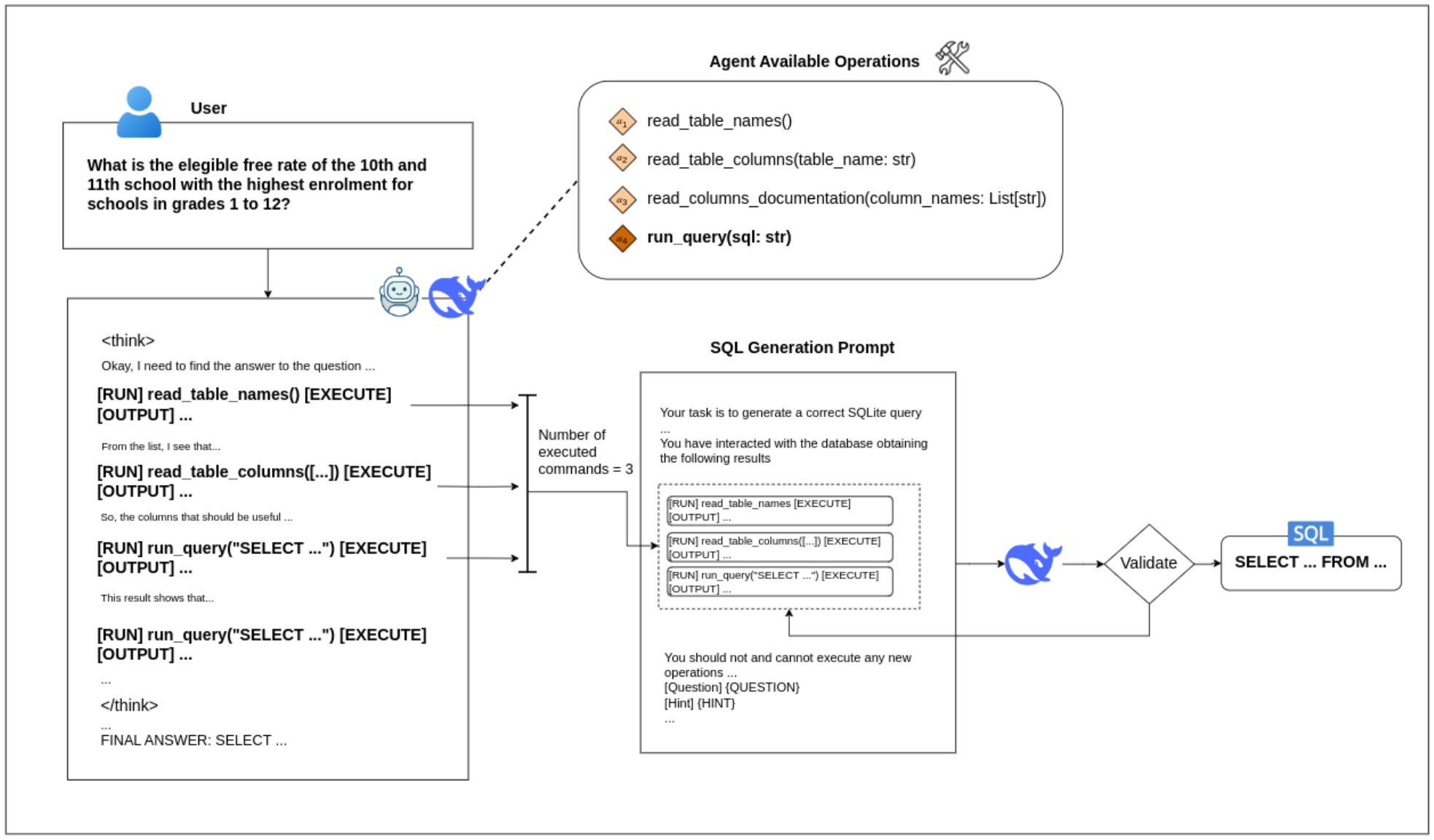}
    \caption{Overview of the framework developed to evaluate the impact of scaling the depth of database understanding on Execution Accuracy (EX). The final SQL generation is created from a prompt containing the commands executed by the agent. We vary the number of commands put into the prompt to evaluate the impact of increasing the depth of the exploration.}
    \label{isolated_db_exploration_pipeline}
\end{figure}

In our experiments, we aim to answer the following research questions:

\begin{enumerate}[1.]
\item What is the impact of adding flexible and dynamic database exploration capabilities to the agent?

\item How does the performance vary by scaling the depth of the exploration performed at inference time?

\item What is the upper bound performance a simple pipeline based on our reasoning agent, integrated with elements to increase diversity, can achieve?

\end{enumerate}

We perform our experiments on a stratified sample of 10\% of the BIRD development set. This dataset is known for covering a diverse set of domains and presenting real-world challenges, such as data cleansing issues and question ambiguities.

To answer questions 1 and 2, we designed a framework to isolate the impact of the dynamic database exploration from any other aspect. Presented in Figure~\ref{isolated_db_exploration_pipeline}, the pipeline is separated into two phases: exploring the database and generating the final answer based on the exploration results.

The database exploration is performed by two agents, the Interaction Agent and the Static Agent. The Interaction Agent is exactly as described in section~\ref{thought_based_planning}, while the Static Agent is the same, except for not having the \textquote{run\_query} operation. This means it can read table names, columns, and documentation, but cannot execute queries arbitrarily to explore the database. We use both agents to generate an answer for the question and keep track of the operations they run in the process.

For generating the final answer, we discard the answers generated by the agents and plug only the operation inputs and outputs into a new prompt to generate the final SQL query. In this way, we can control the number of operations used to generate the final result, making it possible to evaluate its scaling properties, as well as having the setup for final generation identical for both the Exploration Agent and the Static Agent. The only part where they differ is in the operation results plugged into the final prompt.

Also, we evaluate the addition of query refinement in both agents in the final generation step. This step retries the final SQL generation up to 5 times to get a query that executes without errors and without an empty result; at each retry, the previous attempts' error messages are appended to the operations plugged into the prompt. Although this can be seen as part of the dynamic database exploration, we choose to evaluate its impact on the static agent to see how much of the difference between the two agents comes from iterative query refinement and what comes from other aspects of database understanding (e.g., refining the interpretation of the question by examining the data available).

\begin{figure}[h]
    \centering
    \includegraphics[width=1\textwidth]{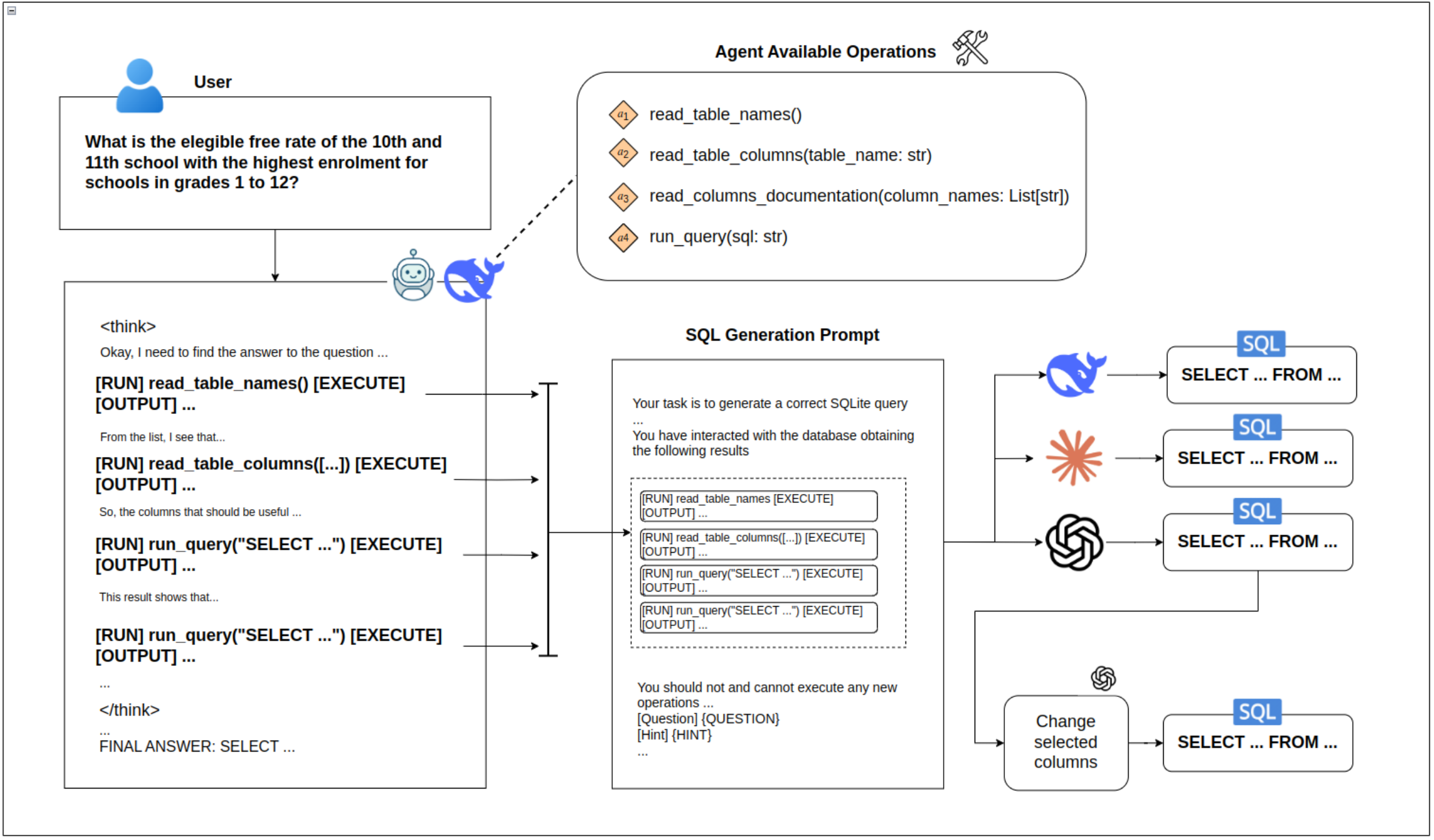}
    \caption{Overview of the full RAISE pipeline. It includes an agent that interacts with the database, generating a series of executed commands. These commands are incorporated into a prompt used to effectively generate SQL queries. To increase diversity, we include two other models, apart from DeepSeek-R1-Distill-Llama-70B: o3-mini and Claude 3.7 Sonnet. The output of o3-mini undergoes a post-processing step that modifies the selected columns without altering the query logic. We also choose to discard the agent's final output as we find that generating the final answer in the subsequent prompt works better, even with the same model.}
    \label{full_pipeline}
\end{figure}

To answer question 3, we use the full RAISE pipeline described in section~\ref{thought_based_planning}, shown in Figure~\ref{full_pipeline}.

\section{Results}\label{results}

\begin{figure}[h]
    \centering
    \includegraphics[width=0.6\textwidth]{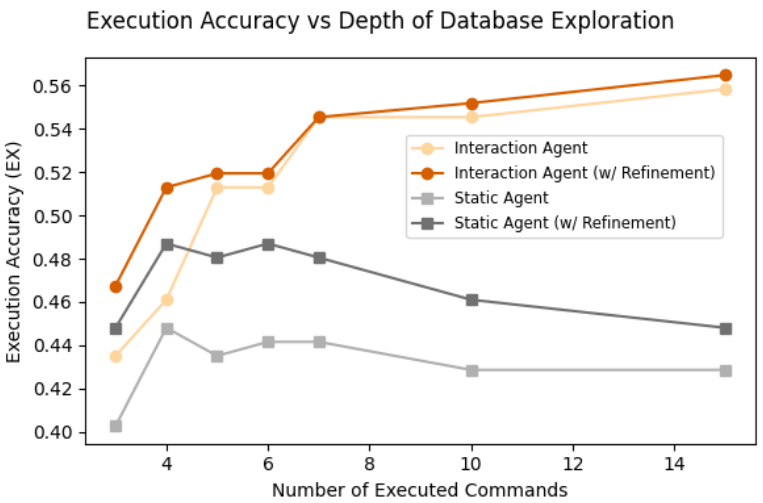}%
    \hspace{3em} 
    \caption{Impact of adding database exploration capabilities to the agent and of scaling the depth of the exploration, measured by the number of executed commands, in Execution Accuracy. Query refinement can be considered part of the exploration process, but we choose to evaluate it separately.}
    \label{test_time_graph}
\end{figure}

\subsection{The Impact of Dynamic Database Exploration}\label{impact_db_exploration}

Figure~\ref{test_time_graph} shows the results for the evaluation setup detailed in Figure~\ref{isolated_db_exploration_pipeline}. We can see that the addition of query refinement improves the Static Agent Execution Accuracy from 42.9\% to 44.8\%. While significant, this improvement is only a fraction of the improvement brought by adding dynamic database exploration capabilities, which elevated the accuracy up to 56.5\%. This result highlights that the gains from flexibly exploring the database at inference time, understanding the nuances of the data and the question, largely exceed the gains of a mere query refinement step.

In addition, it also shows that there is a positive correlation between the depth of the database test-time exploration, measured by the number of commands plugged into the inference prompt, and the Execution Accuracy. The accuracy improves significantly from 3 to 15 commands, showing that the model obtains valuable knowledge to answer the question. However, the accuracy plateaus after 15 commands. This is primarily because the model typically generates fewer than 15 commands. As a result, there are no additional commands to incorporate into the final generation, even if the limit permits it. 

\subsection{Evaluating the Final Pipeline}\label{impact_db_exploration}

\begin{table}[h]
\caption{Upper limit Execution Accuracy of RAISE compared to the candidate generation pipelines of top-performing solutions on the BIRD dataset.}
\label{external_comparison}
\begin{tabular}{@{}l c@{}}
\toprule
Method & \makecell{Execution Accuracy\\(Best-of-N)\footnotemark[1]} \\
\midrule
CHESS & 71.0\% \\
\midrule
CHASE-SQL - Divide and Conquer & 76.0\%  \\
CHASE-SQL - Query Plan & 76.0\%  \\
CHASE-SQL - Online Synthetic Examples & 72.5\%  \\
CHASE-SQL - Full & 82.79\%  \\
\midrule
Reasoning-SQL - Supervised Fine Tuning & 76.5\% \\
Reasoning-SQL - GRPO & 73.5\% \\
\midrule
\textbf{RAISE (ours)} & \textbf{81.8\%} \\
\bottomrule
\end{tabular}
\footnotetext[1]{Except for CHASE-SQL - Full, results are taken from the published scaling graphs, for the largest N available, where the value is close to saturation.}
\end{table}

To assess the full RAISE pipeline’s performance as a drop-in candidate generator for downstream selection, we compare its candidate pool coverage against leading multi-stage systems. Table~\ref{external_comparison} reports the Best-of-N metric, defined as the fraction of questions for which at least one generated candidate matches the gold SQL exactly under BIRD’s execution‐based evaluation. This number represents the upper-bound performance a downstream selection process can achieve when using the generation pipeline.

Across our stratified sample of the BIRD dev set, RAISE achieves a Best-of-N Execution Accuracy of 81.8\%, meaning that the correct SQL query appears among the aggregated candidates from DeepSeek-R1, o3-mini, and Claude 3.7 Sonnet (including the column-order post-processing). This high coverage demonstrates that our unified exploration agent plus diversity generation strategy reliably surfaces correct solutions without any fine-tuning.

\begin{figure}[h]
    \centering
    \includegraphics[width=0.9\textwidth]{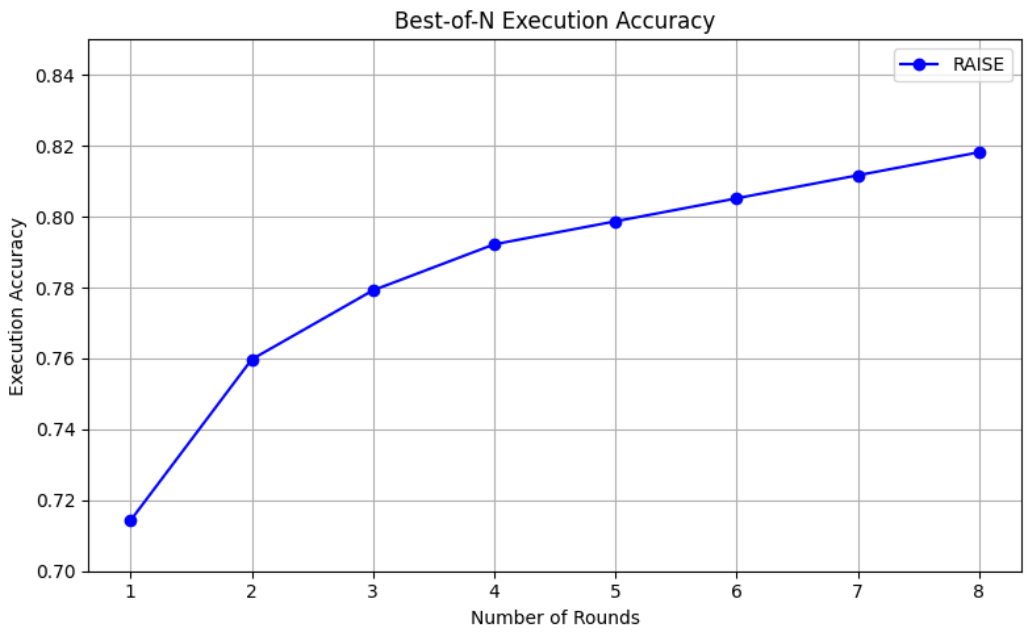}%
    \hspace{3em} 
    \caption{Proportion of questions with at least one correct solution among the candidates on the BIRD dataset.}
    \label{best_of_n_2}
\end{figure}

RAISE’s Best-of-N accuracy of 81.8\% places it among the most competitive candidate-generation pipelines on BIRD (Table \ref{external_comparison}). It outperforms CHESS (71.0\%) by a comfortable margin and all three individual CHASE-SQL variants — Divide and Conquer and Query Plan (both 76.0\%), and Online Synthetic Examples (72.5\%). It also surpasses the two Reasoning-SQL approaches (76.5\% for supervised fine-tuning, 73.5\% for GRPO), despite those being tuned directly on BIRD gold queries. The only pipeline that edges out RAISE in pure best-of-N coverage is CHASE-SQL - Full (82.79\%), which reaches that mark by generating 21 candidates using three specialized generation agents and extensive schema linking and value selection to simplify the database representation. Notably, CHASE-SQL underpins the current top-performing solution on the BIRD benchmark.

Figure~\ref{best_of_n_2} shows the scaling properties of RAISE’s Best-of-N accuracy, with performance steadily improving from 1 to 8 rounds of generation. This consistent gain highlights the effectiveness of the agent in independently exploring and identifying relevant information in the database and in the pipeline producing diverse and relevant solutions.

\section{Conclusion}\label{conclusion}

Text-to-SQL systems have traditionally relied on complex multi-stage pipelines that separate schema linking, query generation, and result validation into specialized modules. Despite advances in large language models, these approaches increase engineering complexity while potentially limiting the natural reasoning capabilities of modern LLMs.

This paper introduces a novel agentic framework that unifies schema linking, query generation, and refinement in a single component. Our agent leverages the internal reasoning traces of LLMs to operate tools to explore the dataset and its documentation, allowing it to interact with the database similar to how humans do. This framework introduces a new strategy for test-time compute scaling in Text-to-SQL, namely, scaling the amount of exploration performed in the database to gain a better grasp of what is stored, how it is stored, and its inconsistencies. Our evaluation shows that such an agent, equipped with strategies to improve diversity, can achieve a best-of-N accuracy of 81.8\% on the BIRD dataset, rivaling top-performing generation components while significantly reducing the engineering complexity.

These findings suggest a promising alternative paradigm for building natural language interfaces to databases that more closely resembles human analytical processes.


While RAISE demonstrates the feasibility and effectiveness of unifying schema linking, query composition, and refinement within a single LLM-based agent, there remain several promising directions to further enhance its capabilities and broaden its applicability. One promising direction is to train the Interaction Agent with reinforcement learning. We hypothesize that the model, when trained to reach the correct answer, will learn to be more effective in exploring the database, increasing the number of executed commands to clarify ambiguities and refine its understanding of the data and the question. 

Another key challenge left for future work is the selection of a single, final SQL query. Because this task is very sensitive to benchmark preferences (e.g. the order the columns are selected), it may not fully capture true answer quality. Future research could fine-tune a dedicated selection model on the BIRD dataset to apply to the RAISE pipeline and develop evaluation metrics more robust to preferences.






\bibliography{sn-bibliography}

\end{document}